%% file: main.tex
\documentclass[10pt,twocolumn,letterpaper]{article}

\include{macros/macros}

\usepackage[pagenumbers]{cvprtemplate/cvpr} 

\usepackage{graphicx}
\usepackage{amsmath}
\usepackage{amssymb}
\usepackage{booktabs}
\usepackage{enumitem}

\definecolor{citecolor}{HTML}{0071bc}
\usepackage[breaklinks=true,bookmarks=false,colorlinks,bookmarks=false, citecolor=citecolor, pagebackref]{hyperref}

\usepackage[capitalize]{cleveref}
\crefname{section}{Sec.}{Secs.}
\Crefname{section}{Section}{Sections}
\Crefname{table}{Table}{Tables}
\crefname{table}{Tab.}{Tabs.}

\def\cvprPaperID{11301} 
\def\confName{CVPR}
\def\confYear{2023}

\begin{document}

\title{A Simple Recipe for Competitive Low-compute Self supervised Vision Models}

\author{Quentin Duval\\
Meta AI\\
{\tt\small qduval@meta.com}
\and
Ishan Misra\\
Meta AI\\
{\tt\small imisra@meta.com}
\and
Nicolas Ballas\\
Meta AI\\
{\tt\small ballasn@meta.com}
}
\maketitle

\begin{abstract}

Self-supervised methods in vision have been mostly focused on large architectures as they seem to suffer from a significant performance drop for smaller architectures. In this paper, we propose a simple self-supervised distillation technique that can train high performance low-compute neural networks. Our main insight is that existing joint-embedding based SSL methods can be repurposed for knowledge distillation from a large self-supervised teacher to a small student model. Thus, we call our method Replace one Branch (RoB) as it simply replaces one branch of the joint-embedding training with a large teacher model. RoB is widely applicable to a number of architectures such as small ResNets, MobileNets and ViT, and pretrained models such as DINO, SwAV or iBOT. When pretraining on the ImageNet dataset, RoB yields models that compete with supervised knowledge distillation. When applied to MSN, RoB produces students with strong semi-supervised capabilities. Finally, our best ViT-Tiny models improve over prior SSL state-of-the-art on ImageNet by $2.3\%$ and are on par or better than a supervised distilled DeiT on five downstream transfer tasks (iNaturalist, CIFAR, Clevr/Count, Clevr/Dist and Places). We hope RoB enables practical self-supervision at smaller scale.

\end{abstract}

\section{Introduction}
\label{sec:intro}

\begin{figure}[h]
    \centering
    \includegraphics[width=\linewidth]{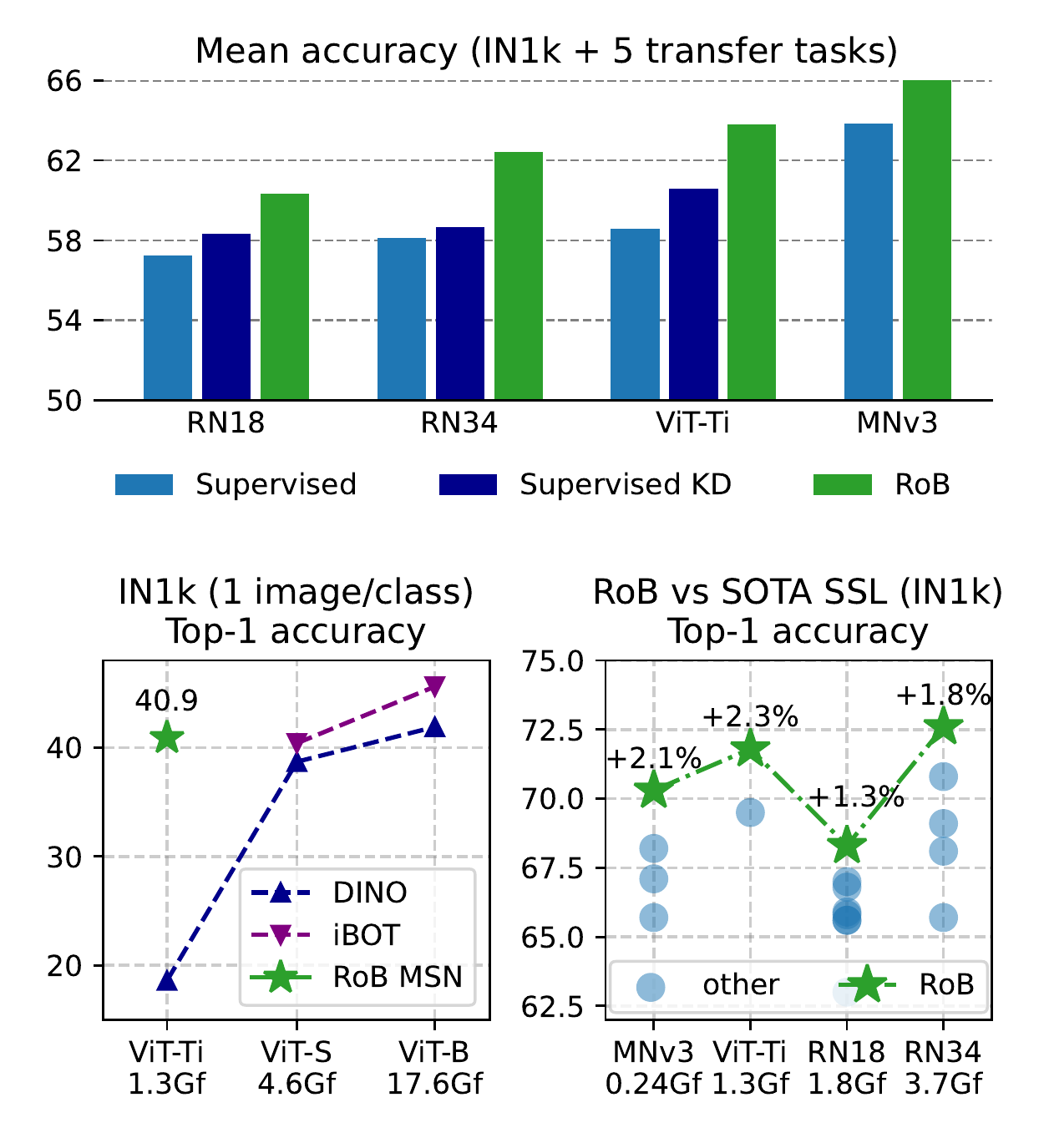}
    \caption{\textbf{Top row}: when averaging performance on 6 tasks (ImageNet-1k and 5 transfer tasks) self-supervised distillation surpasses supervised pretraining and is competitive with supervised distillation. \textbf{Bottom left}: self-supervised distillation of MSN produces students with low-shot performance competitive with DINO or iBOT on larger architectures (measured on ImageNet-1k). \textbf{Bottom right}: RoB compared to previous works on SSL for low-compute network, leveraging knowledge distillation or not.}
    \label{fig:pull_figure}
\end{figure}

\begin{figure*}[h]
    \centering
    \includegraphics[width=\linewidth]{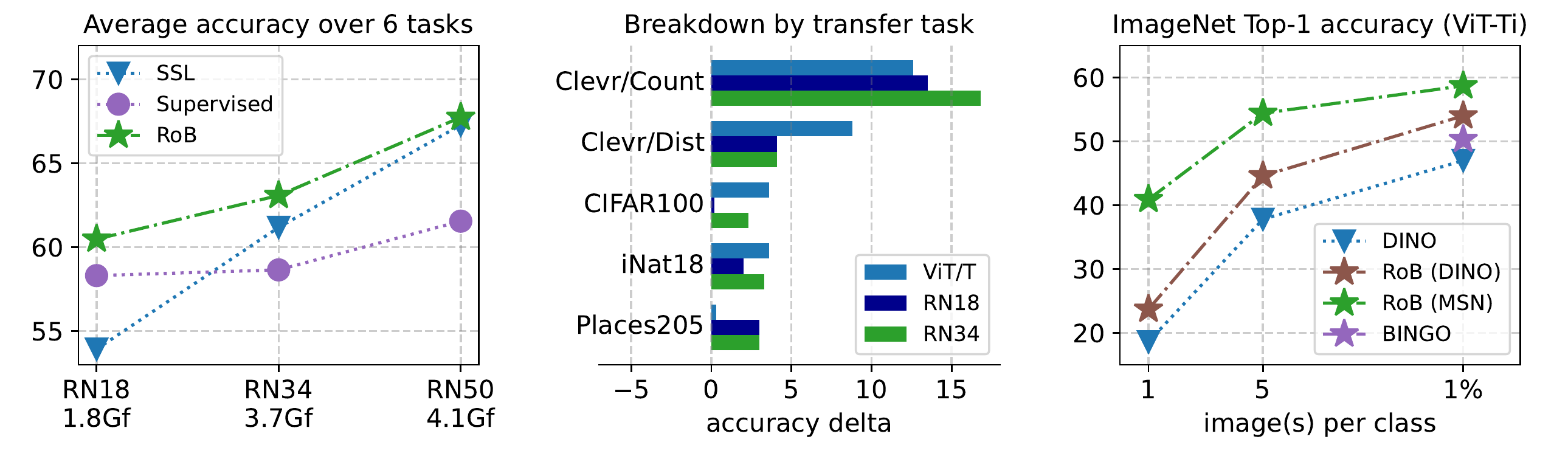}
    \caption{\textbf{Left}: on average over 6 tasks, RoB self-supervised distillation is advantageous over supervised distillation and SSL pretraining without distillation. \textbf{Middle}: Breakdown of transfer performance of RoB against supervised pretraining. \textbf{Right}: RoB ViT-Ti student low-shot performance against DINO ViT-Ti and the self-supervised distillation approach BINGO~\cite{xu2022bag}, SOTA on IN1k $1\%$ for RN18.}
    \label{fig:pull_figure_radar}
\end{figure*}



Self-Supervised Learning (SSL) has demonstrated promising performance for transfer learning~\cite{chen2020simple,caron2021emerging,zhou2021ibotyes,goyal2022vision}. Most methods focus on large neural network architectures with at least  $20$M parameters, ResNet50 architecture and bigger~\cite{he2019moco,caron2020unsupervised,chen2020simple}
or ViT-Small and bigger~\cite{caron2021emerging,zhou2021ibotyes,assran2022masked}
, with the most interesting results obtained with architectures such as ResNet50x4 ($64$GFlops) for SimCLR~\cite{chen2020simple}, or ViT-S/8 ($22$GFlops) and ViT-B/8 ($78$GFlops) for DINO~\cite{caron2021emerging}. At these scales, self-supervised methods have shown stronger transfer and semi-supervised performance than supervised pretraining.
At smaller scales however, ResNet50 and smaller, like ViT-Tiny ($1.3$GFlops) or ResNet18 ($1.8$GFlops) or MobileNetV3~\cite{mobilenet} ($0.24$GFlops), SSL approaches struggle to be competitive with their supervised counterparts~\cite{compress,seed}. This limits the scale of application of SSL methods, as ResNet50 and smaller architectures are the most commonly used architectures by practitioners \cite{https://doi.org/10.48550/arxiv.2106.05237}. Additionally, large models require significantly higher memory, compute, and storage requirements, limiting their use on low compute or low power devices, where lightweight models are usually preferable.



In this paper, we bring the benefits of self-supervised learning to neural network architectures with lower computational requirements.
In particular, we focus on strong off-the-shelf performance on transfer tasks, \eg, linear probe, few-shot, and semi-supervised settings.
We show that Knowledge Distillation~\cite{hinton_knowledge_distillation, compress, seed, xu2022bag, DBLP:journals/corr/abs-2201-05131, disco}, originally developed for supervised learning, can be easily adapted to transfer knowledge from a large self-supervised model (the teacher) to a compact smaller model (the student).


Our main insight is that the underlying training in many existing SSL joint-embedding methods~\cite{chen2020simple, caron2020unsupervised, caron2021emerging, zhou2021ibotyes} can be easily repurposed to a knowledge-distillation framework.
We propose a simple and generic self-supervised distillation method, named Replace one Branch (RoB), that adapts the joint-embedding approaches into distillation methods by simply replacing one of the branches with a pretrained self-supervised teacher.
We instantiate our approach for $4$ such methods: DINO \cite{caron2021emerging}, SwAV \cite{caron2020unsupervised} and iBOT \cite{zhou2021ibotyes} which are currently state of the art in the linear probe transfer setting, and MSN \cite{assran2022masked} as the current state of the start for low shot in-distribution benchmarks (semi-supervised setting).



When pretrained on ImageNet, RoB surpasses previous work in self-supervised distillation on ViT-Tiny and low compute ResNets (ResNet18 and ResNet34). More importantly, we demonstrate that the distilled students significantly surpass their supervised pre-trained counterparts on transfer tasks. More importantly, the self-supervised models are even competitive with low-compute neural nets obtained via supervised distillation (see Figures~\ref{fig:pull_figure} and~\ref{fig:pull_figure_radar}).

\par \noindent \textbf{Contributions}:
\begin{itemize}[leftmargin=*]
    \setlength\itemsep{-0.4em}
  \item We focus on SSL performance using small models and show that existing joint embedding based SSL methods can be repurposed to a knowledge distillation framework.
  \item We propose, RoB, a simple and generic self-supervised distillation approach for joint embedding self-supervised methods. RoB is applicable to a wide variety of methods and architectures.
  \item We demonstrate state of the art self-supervised ViT-Tiny, ResNet18 and ResNet34, improving upon previous work on self-supervision for low-compute networks.
  \item We demonstrate that self-supervision is competitive on linear transfer with supervised pre-training and supervised distillation.
\end{itemize}

\begin{figure*}[h]
    \centering
    \includegraphics[width=\linewidth]{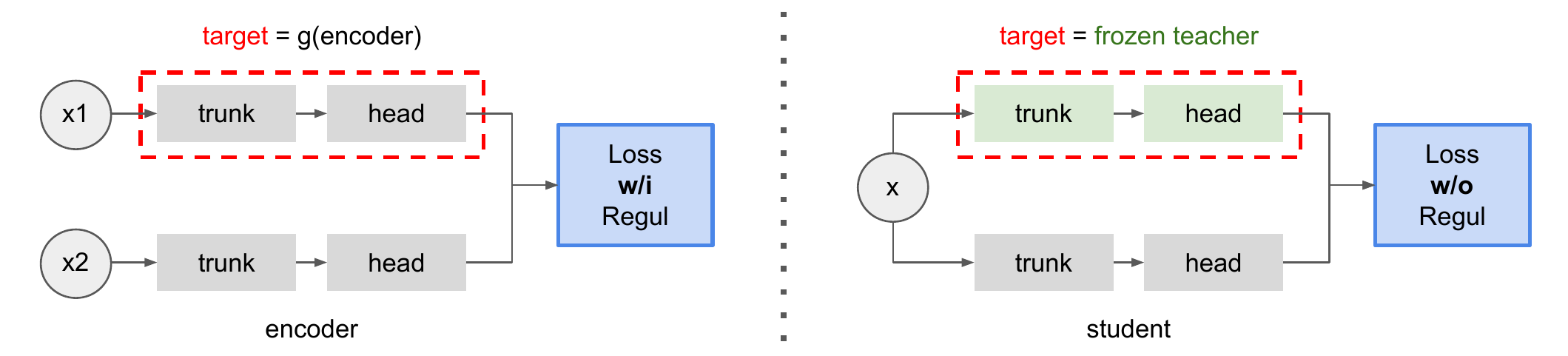}
    \caption{\textbf{Overview of our method.} RoB adapts existing joint-embedding approaches (left) by replacing the branch that produces the target with a frozen teacher pre-trained with the same SSL method (right). Importantly, RoB removes the regularisation terms that aim at preventing collapse from the loss, and use identical-view predictions instead of cross-view predictions in the loss.}
    \label{fig:method}
\end{figure*}

\section{Related work}

\paragraph{Supervised distillation.}
The work of \cite{hinton_knowledge_distillation} introduced the concept of distillation of knowledge from a strong teacher or an ensemble of teacher to a lower-compute student by using a Kullback Leibler (KL) divergence term in the loss. Since then, several works have tried to push the performance of distillation by introducing additional loss terms \cite{https://doi.org/10.48550/arxiv.1412.6550, Tian2020Contrastive, https://doi.org/10.48550/arxiv.2205.10536} or by refining training recipes relying on KL divergence \cite{https://doi.org/10.48550/arxiv.2106.05237}.

Of particular interest to us is DIST \cite{Tian2020Contrastive}, a recent work that produced low compute ResNet18 and ResNet34 from ResNet Strikes Back teachers \cite{https://doi.org/10.48550/arxiv.2110.00476}, reaching state of the art ImageNet-1k top-1 accuracy of $74.5\%$ and $77.8\%$ respectively, and DeiT \cite{touvron2021training} for Vision Transformer. We will use those works as baselines.

\paragraph{Self-supervised Learning.}
While approaches based on reconstruction \cite{bao2021beit, he2021masked} have had a resurgence in the last years, the current state of the art in transfer linear probe is still held by joint-embedding approaches such as DINO \cite{caron2021emerging} or iBOT \cite{zhou2021ibotyes}. Those methods learn to maximize agreement between the representations of two different views of the same image or concept, while making sure that representations do not collapse to the trivial constant solution. Different solutions have been proposed to deal with the collapse issue, contrastive losses \cite{chen2020simple, he2019moco}, soft clustering \cite{caron2020unsupervised, zhou2021ibotyes, caron2021emerging, assran2022masked} and more \cite{https://doi.org/10.48550/arxiv.2006.07733, zbontar2021barlow, bardes2021vicreg}. Our work will concentrate on soft-clustering methods as they currently provide the state of the art linear probe performance.

\paragraph{Self-supervision on low-compute networks.}
Multiple works, such as CompRess \cite{compress}, SEED \cite{seed}, BINGO \cite{xu2022bag}, DisCo \cite{disco} or SimReg \cite{DBLP:journals/corr/abs-2201-05131}, have noticed that joint-embedding self-supervised learning methods such as SwAV \cite{caron2020unsupervised}, MoCo \cite{he2019moco, chen2020mocov2} or DINO \cite{caron2021emerging} suffer from a drop in performance when applied on low compute neural nets. These works have proposed to use Knowledge Distillation \cite{hinton_knowledge_distillation} to circumvent those difficulties. CompRess \cite{compress} and SEED \cite{seed} use a memory queue like MoCo \cite{he2019moco} to distill the knowledge of the teacher by minimizing the cross-entropy between the probability distribution of the teacher and student obtained by comparing a sample to each point in the queue. DisCo \cite{disco} and BINGO \cite{xu2022bag} makes use of contrastive learning, with BINGO additionally grouping samples into cluster of related samples. Finally, SimReg \cite{DBLP:journals/corr/abs-2201-05131} proposes regression as a generic way to transfer feature representation from a teacher to a student.

Another recent line of work \cite{anonymous2023effective, https://doi.org/10.48550/arxiv.2107.14762} has started to explore the reasons behind the performance drop on low compute architectures, with the work~\cite{anonymous2023effective} addressing some of the gap without the recourse to knowledge distillation.

In this work, we will introduce a self-supervised distillation method and show it outperforms those previous approaches for training low-compute networks. Additionally, we will demonstrate that our approach is competitive with supervised training and supervised distillation.



\section{Replace one Branch}
\label{sec:background}
\label{sec:approach}

\paragraph{Background} Joint-embedding architectures~\cite{bromley1993signature} (Figure~\ref{fig:method}, left) aim at learning encoders that produce similar embeddings for different views extracted from a same image.

We consider two views $\mathbf{x_1}$ and $\mathbf{x_2}$ of a given image ${\bf x}$. A student  encoder $f_\theta(\cdot)$ and a teacher encoder $g_\phi(.)$ independently process one view and output the representations $z^S_1$ and $z^T_2$.
Training occurs by pushing the representations $z^S_1$ and $z^T_2$ to be close in order to learn view-invariant teacher/student encoders. The encoder architecture is usually shared across the branches. Teacher weights, however, are not necessary learned and might be updated through of exponential moving average of the student weights~\cite{grill2020bootstrap, caron2021emerging}.

One difficulty with joint-embedding architectures is to prevent representation collapse in which the encoder ignores its inputs and produces a constant image embedding. Several approaches have been investigated in the literature to avoid collapse~\cite{chen2020simple,bardes2021vicreg,caron2020unsupervised}. In this work, we apply our approach on cluster-based methods that use variants of  entropy maximization / prediction sharpening to avoid collapse~\cite{caron2021emerging,assran2022masked}.

\paragraph{RoB Overview}
Replace one Branch (Figure~\ref{fig:method}, left) is a method that explores self-supervised distillation to train low-compute neural network in an unsupervised fashion. 

Replace one Branch (RoB) simple recipe follows two principles: 1) we  replace the teacher encoder $g_\phi$ by an already trained self-supervised teacher $f^T_{\theta^T}$, which is kept frozen during distillation 2) we remove the mechanism that avoids collapse from the self-supervised loss. Representation collapse is indeed not an issue with RoB as the self-supervised teacher is fixed. As a final twist, RoB uses identical-view predictions, where the student representation $z^S_i$ is pushed toward the teacher representation of the same view $z^T_i$.


In the experiment section, we will demonstrate that despite its simplicity, RoB obtains state-of-art SSL performances for low-compute neural network.

We now show how to instantiate RoB for different SSL approaches: DINO~\cite{caron2021emerging}, IBOT~\cite{zhou2021ibotyes}, SwAV~\cite{caron2020unsupervised} and MSN~\cite{assran2022masked}.

\paragraph{RoB-DINO}
To distill from DINO, we replace its  online teacher with an pretrained teacher and remove the centering and sharpening terms in the loss, used to avoid representation collapse.
The RoB loss for DINO becomes:
\begin{equation}
    \mathcal{L_\text{DINO}} = \frac{1}{N} \big( \sum_{i=0}^1 H(z_i^T, z_i^S) + \sum_{i=2}^{N-1} \sum_{j=0}^1 H(z_j^T, z_i^S) \big),
\end{equation}
\label{eq:dino}where $z_i^T$ and  $z_i^S$  are the $i$th teacher and student representations of the $i$-view, and $H$ is the cross entropy. DINO uses  $N$ views from a given images, two large views and $N-2$ small views following muti-crop~\cite{caron2020unsupervised}. RoB uses the same number of views, keeping the same multi-crop mechanism.

\paragraph{RoB-MSN}
MSN~\cite{assran2022masked} tries to match the representation of a masked view, e.g. where some of the patches are randomly dropped,  to an unmasked view. The distillation of MSN is identical to the one of DINO and  only differs in the masking augmentation used in the student encoder.

\paragraph{RoB-iBOT}
iBOT\cite{zhou2021ibotyes} adds a patch-based loss to the DINO loss. The student predicts the representation of masked patches and matches them to the corresponding teacher outputs. We denote by $z_i,p$ the representation of a given patch computed from the view $x_i$. The RoB loss becomes:
\begin{equation}
    \mathcal{L_\text{iBOT}} = \lambda_1 \mathcal{L_\text{DINO}} + \frac{\lambda_2}{2 N_\text{mask}} \sum_{i=0}^1 \sum_{p=1}^{N_\text{mask}} H(z_{i,p}^T, z_{i,p}^S),
\end{equation}
where $z_{i,p}^T$ and $z_{i,p}^S$ denotes respectively the teacher representation and student prediction of the masked patch $p$.


\paragraph{RoB-SwAV}

SwAV relies on Sinkhorn-Knopp (SHK) regularization to avoid representation collapse. With the removal of Sinkhorn-Knopp, the SwAV self-supervised distillation loss $\mathcal{L}_\text{SwAV}$ is the same as the DINO distillation loss.

\section{Experiments}
\label{sec:experiments}

\input{tables/vit_tiny}

\subsection{Experimental setup}

\paragraph{Evaluation datasets.}
We evaluate performance of all of our models with linear evaluation on a range of downstream tasks requiring different levels of abstraction, i.e., classification with CIFAR100~\cite{krizhevsky2009learning}, Places205~\cite{zhou2014learning}, and iNat18~\cite{van2018inaturalist}, with respectively $100$, $205$ and $8142$ categories; object counting with Clevr/Count~\cite{johnson2017clevr, https://doi.org/10.48550/arxiv.1910.04867}; and depth prediction with Clevr/Dist~\cite{johnson2017clevr, https://doi.org/10.48550/arxiv.1910.04867}.
We also evaluate the in-distribution top-1 accuracy on ImageNet-1k ~\cite{russakovsky2015imagenet} (IN1k).

\paragraph{Evaluation protocol.}
For ViT-Tiny, the linear evaluation is performed using either the concatenation of  class tokens extracted from the last 4 layers, as done in DINO \cite{caron2021emerging}, or using only  the last class token, depending on what work best. For ResNets, linear evaluation is performed using the SEER recipes \cite{https://doi.org/10.48550/arxiv.2103.01988} that are available in the VISSL library \cite{goyal2021vissl}. Additional details on those evaluations are provided in the appendix \ref{appendix:eval_details}.




\paragraph{Supervised distillation baselines.}
To show that self-supervised distillation is of any practical interest, we compare it to supervised distillation.

In our experiments, we use DeiT \cite{touvron2021training} and DIST \cite{https://doi.org/10.48550/arxiv.2205.10536} as our supervised distillation baseline. DeiT \cite{touvron2021training} trains a Vision Transformer with an additional distillation token to learn from a RegNet16Gf teacher (with $82.9\%$ top-1 accuracy on ImageNet). DIST \cite{https://doi.org/10.48550/arxiv.2205.10536} distills ResNet Strikes Back \cite{https://doi.org/10.48550/arxiv.2110.00476} to ResNet18 and ResNet34, reaching $74.3\%$ and $77.8\%$ top-1 accuracy on ImageNet respectively.




\paragraph{Teachers.}
Our list of teachers for supervised distillation and self-supervised distillation include the RegNet16Gf trained in DEIT \cite{touvron2021training}, the ResNet50 trained in ResNet Strikes Back \cite{https://doi.org/10.48550/arxiv.2110.00476}, the SSL  ResNet50 and RegNet128Gf\cite{caron2020unsupervised, https://doi.org/10.48550/arxiv.2103.01988} and the iBOT \cite{zhou2021ibotyes}, DINO \cite{caron2021emerging} and MSN \cite{assran2022masked} ViT models. All teachers and their top-1 accuracy on ImageNet are listed in Appendix \ref{appendix:teachers_list}.

\subsection{Transfer performance of SSL distillation}
\label{sec:vit_tiny}

In this section, we investigate the transfer performance of self-supervised distillation and demonstrate RoB is a viable alternative to supervised distillation.

\paragraph{Vision Transformer.} We distill DINO \cite{caron2021emerging} and iBOT \cite{zhou2021ibotyes} teachers to a vanilla ViT-Tiny with 16x16 patches. Specifically, we rely on three different teachers: a ViT-S/8 pretained with DINO, a ViT-B/16 and a ViT-L/16 both pretrained with iBOT. Our students are trained using the recipe described in section \ref{sec:approach} for 300 epochs. For both method, the student uses a projector head with a similar architecture than the teacher head, but changing its input dimension to $192$ to match the student encoder output dimension.

We compare our results with DeiT~\cite{touvron2021training} . Specifically, we use two supervised DeiT ViT-Tiny baselines, trained with and without knowledge distillation and reaching respectively  $72.2\%$ and $74.5\%$ ImageNet-1k top-1 accuracy. 
From the results in Table~\ref{tab:vit_tiny}, we observe that 1) self-supervised distillation has better linear probe performances on transfer tasks compared to supervised pre-training. RoB is also competitive on ImageNet-1k 2) RoB outperforms DeiT with supervised distillation on most transfer tasks 3) self-supervised distillation achieves better linear probe performances with stronger teachers. An iBOT-L/16 teacher leading to strictly superior performance on all benchmarks when compared to an iBOT-B/16.

RoB achieves $71.8\%$ with a ViT-Tiny on ImageNet-1k which outperforms by a significant $2.3\%$ the best reported performance by self-supervised method on this architecture~\cite{anonymous2023effective}.

\paragraph{ConvNets.} We also reports result on ResNet-18 and ResNet-34 architecture in Table~\ref{tab:swav_mean} to demonstrate the versability of RoB. We compare RoB with two supervised pre-trained baselines: torchvision \cite{marcel2010torchvision} and ResNet Strikes Back (RNSB) \cite{DBLP:journals/corr/abs-2110-00476}. For RNSB, we use the $A1$ training recipe. We also use DIST \cite{https://doi.org/10.48550/arxiv.2205.10536} to train supervised distilled ResNet18 and ResNet34.

Similar to our previous results with ViT-Tiny, RoB models show better transfer performance than their supervised counterparts. RoB is also competitive with supervised distillation on all benchmarks.
We finally explore the distillation of DINO/S8 teacher with a ResNet student. Using a ViT as teacher does improve the average performance on our set of 6 tasks, especially on the ResNet34 where the gap is the most pronounced.
We include the detailed results for ResNets  and additional experimental details in Apppendix~\ref{appendix:training_details} and ~\ref{appendix:full_results}. Results for MobileNetV3 are reported in Appendix~\ref{experiment:mobilenet}.

\subsection{Semi-supervised setting}

In this section, we show that RoB can produce low-compute students with competitive low-shot classification performances.

We explore low-shot classification on Imagenet-1K~\cite{assran2022masked} using 1, 5 or 12/13 labeled images per class. 12/13 images per class corresponds to $1\%$ of  training data. 
We follow a linear evaluation protocol and fit a linear layer on top of a frozen encoder with  logistic regression~\cite{caron2021emerging}. Results reported for $1$ image per class and $5$ images per class settings are averaged over 5 and 3 ImageNet splits respectively. Additional details on evaluations are available in Appendix \ref{appendix:eval_details}.

We use RoB to distill a ViT-B/16 MSN teacher \cite{assran2022masked}, a method achieving state-of-art performances in low-shot classification, to a ViT-Tiny student. We also the use a DINO and a iBOT teachers. We train our students for 300 epochs. For MSN, we use an input masking ratio of  of $5\%$ which we found empirically that this ratio achieves better results than $10\%$ or $0\%$.


Results are shown in Table \ref{tab:msn_low_shot}. We compare our results with the self-supervised distillation approach BINGO \cite{xu2022bag}, DINO and iBOT self-supervised models.
Our MSN student shows competitive low-shot performance and achieves better low-shot performances than BINGO and vanilla ViT-Tiny model trained with DINO. RoB even outperforms ViT-S/16 baselines trained with DINO or iBOT in the case of 1 image per class, an architecture with $4$ times more parameters and $4$ times more FLOPS. 

\input{tables/msn_low_shot}

\subsection{Comparison with the state of the art}

We compare RoB to the self-supervised distillation methods CompRess \cite{compress}, SEED \cite{seed}, BINGO \cite{xu2022bag} SimReg \cite{DBLP:journals/corr/abs-2201-05131} and DisCo \cite{disco} on ResNet18, ResNet34 and MobileNetV3. We focus on the ImageNet-1k linear probing, the common benchmark reported across previous works. Resnet results are available in Table \ref{tab:sota}, MobileNetV3 in Appendix~\ref{experiment:mobilenet}.

We train ResNet18 and ResNet34 students with RoB, first using ResNet50 \cite{caron2020unsupervised} teacher and a  RegNet128Gf teacher \cite{https://doi.org/10.48550/arxiv.2103.01988} both pretrained with SwAV~\cite{caron2020unsupervised}. We also explore the use of a DINO/S8 pretrained teacher with RoB on ResNets. Refer to Appendix \ref{appendix:cross_arch_distillation} for more experimental details.

 Our method outperforms all previous methods on the ResNet18 and ResNet34 architectures, pushing the in-distribution linear probe results by $1.3\%$ for a ResNet18 and by $1.8\%$  on a ResNet34. 
\input{tables/sota}

\input{tables/swav_mean}

\subsection{Beyond low-compute networks}

We demonstrate the RoB can also improve the performance of self-supervised ResNet50 using a Vision Transformers  teacher such as DINO \cite{caron2021emerging}.

We use RoB on a DINO/S8 teacher and distill it to a ResNet50 student. We train our student  for 100 epochs. Results are reported in Table \ref{tab:swav_mean}. The student ResNet50 outperforms the same model trained using DINO  ($+0.65\%$ across our 6 tasks) as well a SwAV ResNet50 ($+0.46\%$ across our 6 tasks) and supervised pretraining ($+6.18\%$ across our 6 tasks). The detailed results are available in Appendix \ref{appendix:full_results} while additional training details on cross-architecture distillation are available in Appendix \ref{appendix:cross_arch_distillation}.

\subsection{Visualizing the transferred knowledge}

\begin{figure*}[t]
    \centering
    \includegraphics[width=\linewidth]{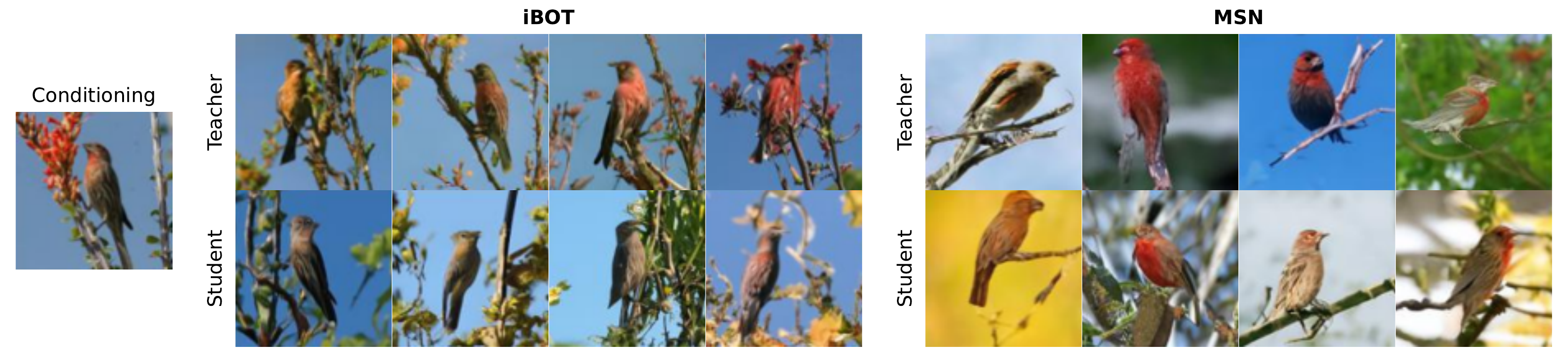}
    \caption{\textbf{RCDM \cite{bordes2022high} visualisations} applied on iBOT (on the left) and MSN (on the right) both teacher (top row) and student (bottom row). We use RCDM to enable visualization of the learned SSL representation. Samples are generated by conditioning on a the left image representation and with various random seeds. Features that remain constant across samples depict information contained in the SSL representation, whereas features that vary depict information that is not contained. We qualitatively observe that the iBOT conserves more information about the context around the bird while MSN abstracts those details away. The RoB-iBOT and RoB-MSN students visually inherits these characteristics from their respective teachers.}
    \label{fig:rcdm}
\end{figure*}

\begin{figure}[h]
    \centering
    \includegraphics[width=\linewidth]{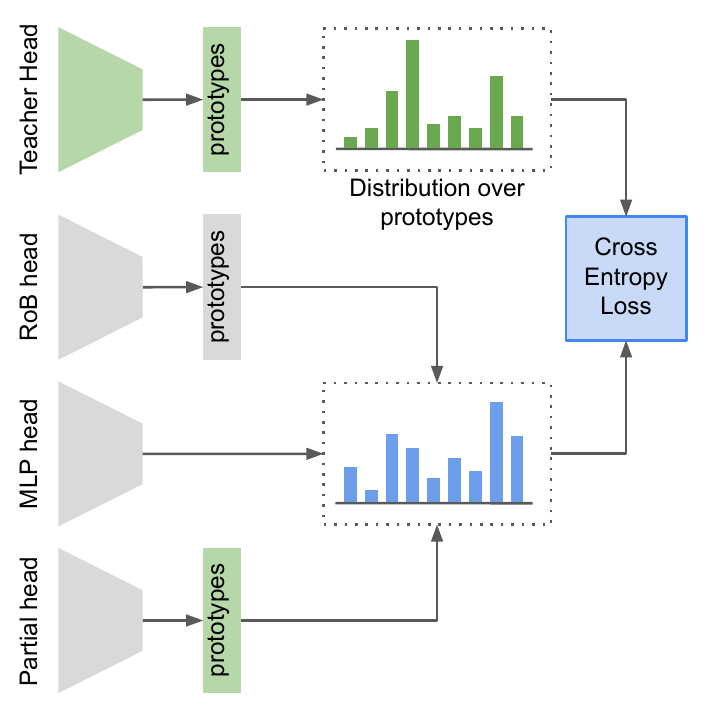}
    \caption{\textbf{Student head designs.} We explore three different kinds of student head designs of joint-embedding prototype based methods. First is RoB default design: student and teacher have different heads. Second is replacing the student head with a MLP. Third is to use the teacher prototypes in the student head.}
    \label{fig:head_designs}
\end{figure}

We now use the RCDM \cite{bordes2022high} conditional diffusion model to visualize the prototypes of the prototypes learned with our RoB students. As highlighted by \cite{bordes2022high}, conditional diffusion models are useful to get a better understanding of what is learned by neural networks. We qualitatively compare the results obtained in the student against the results of the teacher to assess how much knowledge has been transferred.

We compare the representations of both iBOT and MSN teacher against the representations of both RoB-iBOT and RoB-MSN student in Figure \ref{fig:rcdm}. Qualities that remain constant across samples depict information contained in the SSL representation, whereas qualities that vary depict information that is not contained in the representation.
MSN retains less details about background and poses than iBOT, and this difference is kept when we observe the RoB-MSN and RoB-iBOT students. We hypothesize that these difference are connected to the superior performance of RoB-MSN on low-shot in-distribution, where removing background details lessens the risk of over-fitting, while also explaining why RoB-iBOT representations lead to stronger transfer performance where more labels are available.

\subsection{Ablations}


\paragraph{Student head designs.}
Because RoB replaces one branch of the distilled joint-embedding approach with a pre-trained teacher, the design of the projection head of the teacher and the student are identical (Teacher head and RoB head in Figure \ref{fig:head_designs}), only differing on their input dimension. In the case of joint-embedding approaches based on soft clustering \cite{caron2020unsupervised, caron2021emerging,zhou2021ibotyes, assran2022masked}, the projection head is composed of a projection to a low dimension space ($128$ for SwAV), normalized, followed by a dot product with a set of learned prototypes used to compute a distribution over the prototypes. We investigate whether the method can be simplified and the performance improved by using a different projection head for the student.

Our first experiment is to replace the SSL projection head of the student with a simpler multi-layer perceptron head (MLP head in Figure \ref{fig:head_designs}). The output dimension of this prediction head is set to be the number of prototypes of the teacher, $3000$ in the case of SwAV, to directly predict the distribution given by the teacher. We try MLPs of depth $1$, $2$ or $3$ and report the best result (obtained with depth $2$). 


Our second experiment consists in keeping the student projection head intact but using the prototypes of the teacher in the student head instead of the student learning its own set of prototypes. We dubbed this approach Partial head (see Figure \ref{fig:head_designs}) as the projection head is partially learned.





The results are available in Table \ref{tab:ablation_head}. Distilling with a MLP head for the student does improve performance up to $0.3\%$ improvement on ImageNet, but suffers from a up to $4.9\%$ drop in terms of transfer performance on tasks such as CIFAR100 or iNaturalist18, while distilling using the partial head design for the student significantly degrades performance on both on ImageNet and all transfer tasks.

\paragraph{Importance of multi-crop.}
As RoB distills the joint-embedding approach by mimicking it, only replacing one branch with a pretrained teacher and adapting the loss, RoB will make use of multi-crops \cite{caron2020unsupervised} if the method being distilled does use multi-crops. We investigate here if this is necessary, or if the student can be distilled using large crops alone.

We use RoB to distill a DINO/S8 \cite{caron2021emerging} teacher and train for 300 epochs once with and once without multi-crops and compare the results. Table \ref{tab:ablation_views} shows that without multi-crop, performance is decreased significantly on ImageNet with a $1\%$ decrease in top-1 accuracy, as well as on transfer tasks with up to $4.4\%$ accuracy drop on iNaturalist18, the transfer task the most sensitive to multi-crop. This shows the critical importance of multi-crop in the context of self-supervised distillation. Additional detailed results are provided in \ref{appendix:full_results}.

\paragraph{Identical-view vs cross-view predictions.}
As described in Section \ref{sec:approach}, RoB distills joint-embedding approaches by using identical-view predictions, where the student learns to match the teacher representation of the same view, instead of using cross-view predictions, where the student would learn to predict the teacher representation of a different view of the same original image.

Cross-view predictions are used in joint embedding methods to learn invariance to augmentations. However, as the desired invariances are already learned by the teacher, our hypothesis is that identical view predictions will suffice to learn the useful invariances encoded by the teacher while making training more stable, leading to equivalent or slightly better performance on all tasks.

To verify our hypothesis, we distill both a DINO S/8 \cite{caron2021emerging} teacher and a iBOT L/16 \cite{zhou2021ibotyes} teacher with either cross-view predictions or identical-view predictions and compare the results obtained. As shown in Table \ref{tab:ablation_views}, distilling with identical-view prediction yields overall equal or better linear probe results, with differences being more pronounced on iBOT than on DINO. RoB therefore adopts identical-view predictions as part of the method.

\input{tables/ablation_views}
\input{tables/ablation_head}

\section{Limitations and Conclusion}

We explored a simple and generic approach for self-supervised distillation that consists in "Replacing one Branch" (RoB) in joint-embedding SSL methods by a pretrained teacher, while adapting the loss of the method to remove mechanisms preventing representation collapse.

We instantiated RoB on 4 different joint embedding approaches and demonstrated that RoB produces state of the art SSL students with properties that follow those of their respective teachers. RoB applied on SwAV, DINO or iBOT leads to transfer performance that competes with supervised distillation, fixing the performance gap of SSL on low-compute architectures, while RoB-MSN produces students with strong semi-supervised capabilities.


\pagebreak

\section*{Acknowledgements}

We would like to thank Shagun Sodhani for his helpful feedback all along the project and his help in proof reading the paper. We also would like to thank Olivier Delalleau and Mike Rabbat for their helpful feedback and discussions around the subject of SSL and knowledge distillation.

{\small
\bibliographystyle{cvprtemplate/ieee_fullname}
\bibliography{egbib}
}

\clearpage
\vfill
\pagebreak
\appendix

\section{Additional Experiments}

\subsection{Tiny compute experiments}
\label{experiment:mobilenet}

In Section \ref{sec:experiments}, we demonstrated that RoB produces students that transfer better than supervised pretraining and surpasses previous self-supervised methods on low-compute ViT-Tiny and ResNets. In this section, we extend our experiments to MobileNetV3~\cite{mobilenet}, an architecture with one order magnitude less compute than our other student networks (Table \ref{tab:student_list}).

\paragraph{Experiments.}
We distill a ResNet50 pretrained with SwAV~\cite{caron2020unsupervised} to a MobileNetV3~\cite{mobilenet} using the standard protocol of RoB. We use the MobileNetV3 implementation of torchvision~\cite{marcel2010torchvision}. In contrast to ResNets, the MobileNetV3 pretrained on ImageNet has a head composed to two linear layers separated by a Hardswish activation layer. We remove this MobileNetV3 classifier head, and replace it with a standard SwAV head with $3000$ prototypes. We train the student for 200 epochs. Additional details on RoB distillation for MobileNetV3 are provided in Appendix \ref{appendix:training_details}.

\paragraph{Transfer performance.}
We now compare our MobileNetV3 student to Torchvision~\cite{marcel2010torchvision} and timm~\cite{rw2019timm} supervised pretrained models in terms of transfer performance.

Here we depart from the linear evaluation protocol and use the MobileNetV3 head instead of a linear layer as classifier head, to stay consistent with the original MobileNetV3 architecture. The MobileNetV3 head is composed of two linear layers separated with a Hardswish activation layer. More details are available in Appendix~\ref{appendix:eval_details}.

The results are available in Table~\ref{tab:mobile_nets}. RoB produces students that surpass, in average over 6 tasks, the performance of supervised pretraining.

\input{tables/appendix/mobile_net_transfer}

\paragraph{Comparison with the state of the art.}
We compare our MobileNetV3 student with SEED~\cite{seed}, BINGO~\cite{xu2022bag} and DisCo~\cite{disco} on the ImageNet top-1 accuracy. Again, we use the MobileNetV3 head to stay consistent with MobileNetV3~\cite{mobilenet} architecture, instead of a linear layer. Additional details on evaluations are provided in Appendix~\ref{appendix:eval_details}.

We report our results in Table~\ref{tab:mobile_nets_sota}. RoB surpasses other methods by more than $2\%$, despite using a much smaller teacher than the other approaches. When using a linear layer instead of the MobileNetV3 head, RoB also surpasses the state of the art results with $68.6\%$ top-1 accuracy.

\input{tables/appendix/mobile_net_sota}

\subsection{Nearest Neighbors}

Following SEED~\cite{seed}, CompRess~\cite{compress} and BINGO~\cite{xu2022bag}, we report kNN evaluations of all of our students models on ImageNet. We use the kNN implementation of VISSL~\cite{goyal2021vissl}, based on cosine similarity, and report top-1 accuracy for 10 neighbors and 20 neighbors, using the output features of the encoder (the class token for the ViT-Tiny and the output of the average pooling for the ResNet18 and ResNet34).

As shown in Table~\ref{tab:knn_results}, RoB surpasses the other methods by $+2.7\%$ on a ResNet18 and $+3.7\%$ on a ResNet34. Our result on ViT-Tiny, $69.5\%$ is equal to the SOTA SSL results~\cite{anonymous2023effective} reached with linear evaluation, despite using kNN only.

\input{tables/appendix/knn}

\subsection{List of teachers}
\label{appendix:teachers_list}

The full list of teacher used for our experiments is available in Table \ref{tab:teacher_list}. The weights of teachers are downloaded from their respective official repository, with the exception of the SwAV weights, downloaded from VISSL~\cite{goyal2021vissl}, which contains a valid implementation of SwAV.

\input{tables/appendix/teachers_list}

\input{tables/appendix/student_list}

\section{Training details}
\label{appendix:training_details}

We now describe the key ingredients of the training recipes for RoB on DINO, iBOT, SwAV and MSN. All code and configurations needed will be released to make sure our results are reproducible.

\subsection{Rob-DINO}
\label{appendix:training_details_dino}

For RoB-DINO, we use the recipe to train DINO-S/16~\cite{caron2021emerging} as a baseline. We list those details below.

\paragraph{Augmentations.}
We use the default data augmentations of DINO~\cite{caron2021emerging} with 2 large crops and 8 small crops. The teacher only processes the large 2 crops while the student processes all 10 crops, and learn to map the representation of each large crop to its corresponding large crop on the teacher side, and the representation of each small crop to the representation of the large crops of the teacher.

\paragraph{Optimization.}
Our students are trained with AdamW. The learning rate follows a cosine schedule with a 10 epochs linear warm up with a peak learning rate of $2e^{-3}$. We train on 16 GPUs with a total batch size of 1024. The weight decay also follows the same cosine schedule as specified in DINO~\cite{caron2021emerging} increasing from $0.04$ to $0.4$ at the end of training. The student is trained with a drop path rate of $0.1$.

\paragraph{Loss parameters.}
We use the default parameters of DINO~\cite{caron2021emerging}. The teacher softmax temperature is set to $0.07$ and the student softmax temperature to $0.1$. The centering of the original DINO~\cite{caron2021emerging} is removed.

\subsection{Rob-iBOT}

For RoB-iBOT, we use the recipe to train iBOT-S/16~\cite{zhou2021ibotyes} as a baseline. The details are listed below.

\paragraph{Augmentations.}
We use the default data augmentations of iBOT~\cite{caron2021emerging} with 2 large crops and 10 small crops. The teacher only processes the large 2 crops while the student processes all 12 crops, and learn to map the representation of each large crop to its corresponding large crop on the teacher side, and the representation of each small crop to the representation of the large crops of the teacher. The student also learns to map the masked patches representation to the corresponding teacher patches for the large crops.

\paragraph{Optimization.}
Our students are trained with AdamW. The learning rate follows a cosine schedule with a 10 epochs linear warm up with a peak learning rate of $2e^{-3}$. We train on 16 GPUs with a total batch size of 1024. The weight decay follows a cosine schedule increasing from $0.04$ to $0.48$ at the end of training. The student is trained without drop path rate.

\paragraph{Loss parameters.}
The teacher softmax temperature is set to $0.07$ for both patch and class token. The student softmax temperature to $0.1$. We do not use a schedule for the temperature as used in iBOT~\cite{zhou2021ibotyes} as the teacher is kept fixed during training.

\subsection{Rob-SwAV}

For RoB-SwAV, we use the recipe to train a ResNet50 SwAV~\cite{caron2020unsupervised} available in VISSL~\cite{goyal2021vissl} as a baseline. The details are listed below.

\paragraph{Augmentations.}
We use the default data augmentations of SwAV~\cite{caron2020unsupervised} with 2 large crops and 4 small crops. The teacher only processes the large 2 crops while the student processes all 6 crops, and learn to map the representation of each large crop to its corresponding large crop on the teacher side, and the representation of each small crop to the representation of the large crops of the teacher.

\paragraph{Optimization.}
Our students are trained with LARS with momentum $0.9$. The learning rate follows a cosine schedule with a 10 epochs linear warm up with a peak learning rate of $4.8$. We train on 64 GPUs with a total batch size of 4096. The weight decay is set to $1e^{-6}$.

\paragraph{Loss parameters.}
The teacher softmax temperature is set to $0.03$ and the student softmax temperature to $0.1$. We disable Sinkhorn-Knopp on the teacher prototypes as representation collapse is not an issue since the teacher is frozen.

\subsection{Rob-MSN}

Our training recipe is inspired from the MSN~\cite{assran2022masked} training recipe for ViT-S/16. The details are listed below.

\paragraph{Augmentations.}
We use the default data augmentations of MSN~\cite{assran2022masked} with 2 large crops and 10 small crops. The teacher only processes the large 2 crops while the student processes all 12 crops, and learn to map the representation of each large crop to its corresponding large crop on the teacher side, and the representation of each small crop to the representation of the large crops of the teacher. We use the patch dropping mechanism of MSN~\cite{assran2022masked}, dropping $5\%$ of the patches on the student side for ViT-Tiny.

\paragraph{Optimization.}
Our students are trained with AdamW. The learning rate follows a cosine schedule with a 10 epochs linear warm up with a peak learning rate of $2e^{-3}$. We train on 16 GPUs with a total batch size of 1024. The weight decay follows a cosine schedule increasing from $0.04$ to $0.4$ at the end of training. The student is trained without drop path rate as done in typical MSN~\cite{assran2022masked} training.

\paragraph{Loss parameters.}
Both the teacher and student softmax temperature are set to $0.1$. We disable the entropy maximization of MSN~\cite{assran2022masked} as represention collapse is not an issue since the teacher is kept frozen.

\subsection{Cross architecture distillation}
\label{appendix:cross_arch_distillation}

For cross-architecture distillation, we use the same recipe as if the student was of the same architecture. In the case of RoB-DINO from a DINO-S/8 to a ResNet, we therefore use the recipe described in the Appendix \ref{appendix:training_details_dino} above.

\section{Evaluation details}
\label{appendix:eval_details}

We use the default linear evaluation configurations of VISSL~\cite{goyal2021vissl} to evaluate our models. The most important details are listed below.

\subsection{Linear evaluation of ViT-Tiny}
\label{appendix:eval_vit}

For Vision Transformers~\cite{dosovitskiy2020image} models, we report the best linear classifier number among the following representations:
\begin{itemize}
\item the concatenation of the last $4$ layers of the class token
\item the representation of the last layer of the class token.
\end{itemize}

We attach $2$ linear heads per chosen representation, one composed of a single linear layer, and one with an added batch normalization before the linear layer, and report the best result among those two heads.

\subsection{Linear evaluation of ResNets}
\label{appendix:eval_resnets}

For ResNet~\cite{he2016deep} models, we follow the evaluation protocol of SEER~\cite{goyal2022vision} and report the best linear classifier number among the following representations:
\begin{itemize}
\item the final representation layer (of dimension $512$ for ResNet18 and ResNet34, and $2048$ for ResNet50)
\item an adaptive average pooling of the last feature map, concatenated to get a single representation (for example on ImageNet, we use representations of dimension $1024$ for a ResNet18 or ResNet34)
\end{itemize}

We attach $2$ linear heads per chosen representation, one composed of a single linear layer, and one with an added batch normalization before the linear layer, and report the best result among those two heads. 

\subsection{Evaluation of MobileNets}

The MobileNetV3~\cite{mobilenet} classifier head, unlike the typical classification head used for ResNets or ViT, is composed of two linear layers, with a Hardswish activation and a drop out of $20\%$ in between. The equivalent PyTorch~\cite{pytorch} code is:

\begin{verbatim}
head = torch.nn.Sequential(
    torch.nn.Linear(960, 1280),
    torch.nn.Hardswish(inplace=True),
    torch.nn.Dropout(p=0.2, inplace=True),
    torch.nn.Linear(1280, num_classes),
)
\end{verbatim}

To be able to fairly compare our students to supervised pretrained models, we therefore depart from linear evaluation and use the MobileNetV3 head instead the traditional linear head in all of our evaluations.

Finally, to better deal with representations of different norms, we follow Appendix~\ref{appendix:eval_vit} and \ref{appendix:eval_resnets} and add a batch normalization in between the frozen encoder and the head.

\subsection{Low shot protocol}
\label{appendix:low_shot_protocol}

We first extract the features of the encoder at the class token representation for Vision Transformers~\cite{dosovitskiy2020image} and after the average pooling layer for ResNets~\cite{he2016deep}, after resizing and central cropping the images.

We then normalize the features and subtract the mean, using the training set statistics for both train and test sets. We fit a logistic regression classifier on top of those representations. We use the scikit-learn~\cite{scikit-learn} logistic regression implementation, sweep a large range of lambdas from $10^4$ to $10^{-2}$, evenly spaced on a logarithmic scale, and report the best results among those.

We then average our results over multiple splits of ImageNet, $5$ splits for $1$ image per class, $3$ splits for $5$ images per class, and 1 split for ImageNet-$1\%$.

\section{Full Result Tables}
\label{appendix:full_results}

We now list the details of all aggregated metrics reported in the experimental Section~\ref{sec:experiments}.

\input{tables/appendix/resnets_full}

\end{document}

%% file: macros/macros.tex
\usepackage[table,dvipsnames]{xcolor}

\usepackage{color}
\usepackage{wrapfig}
\usepackage{xspace}
\usepackage{bbold}
\usepackage[nice]{nicefrac}
\usepackage{mathtools}
\usepackage{booktabs}
\usepackage{fontawesome}
\usepackage[font={footnotesize}]{caption}
\usepackage{subcaption}
\usepackage{stfloats}
\usepackage{pbox}
\usepackage{multirow}
\usepackage{microtype}

\definecolor{fbApp}{HTML}{c8e7fa}
\definecolor{lightbrown}{HTML}{d8cbc4}
\definecolor{darkgreen}{HTML}{006600}
\newcommand{\cc}{\cellcolor{fbApp}}
\newcommand{\ac}{\cellcolor{lightbrown}}

%% file: tables/vit_tiny.tex
\begin{table*}[]
    \centering
    \begin{tabular}{c|c c c |c c c c c c | c}
        Arch & Method & Teacher & Flops & IN1k & iNat18 & Clevr/C & Clevr/D & Cifar100 & Places205 & Mean \\
        \midrule
        \multicolumn{10}{l}{\small\bf\it Methods without distillation}\\
        ViT/T16 & DEIT & - & - & 72.2 & 36.3 & 67.2 & 54.1 & 73.5 & 48.1 & 58.57 \\
        ViT/T16 & DINO & - & - & 66.2 & 36.7 & 77.5 & 64.8 & 74.9 & 46.4 & 61.08 \\
        \midrule
        \multicolumn{10}{l}{\small\bf\it Supervised distillation}\\
        ViT/T16 & DEIT & RegNet16Gf & 16.0Gf & \bf 74.5 & 38.1 & 70.7 & 56.7 & 74.8 & 48.6 & 60.57 \\
        \midrule
        \multicolumn{10}{l}{\small\bf\it Self-supervised distillation}\\
        ViT/T16 & RoB & DINO S/8 & 22.4Gf & 71.4 & \bf 39.8 & 79.8 & 62.9 & \bf 77.1 & 48.4 & \cc 63.23 \\
        ViT/T16 & RoB & iBOT B/16 & 17.6Gf & 71.4 & 36.9 & 81.3 & 65.1 & 75.7 & 49.1 & \cc 63.25 \\
        ViT/T16 & RoB & iBOT L/16 & 61.6Gf & 71.8 & 37.4 & \bf 81.3 & \bf 66.0 & \bf 76.8 & \bf 49.6 & \cc \bf 63.82 \\
        \bottomrule
    \end{tabular}
    \caption{\textbf{RoB transfer performance on ViT-Tiny}. Distilling DINO or iBOT leads to better  linear probe transfer performances than supervised pre-training, supervised distillation, or self-supervised learning. Importantly, RoB improves with better teachers, with a distilled iBOT-L/16 teacher having better transfer perfformances than iBOT-B/16. The in-distribution performance of RoB students is almost equal to the supervised baseline without distillation.}
    \label{tab:vit_tiny}
\end{table*}

%% file: tables/msn_low_shot.tex
\begin{table}[]
    \centering
    \begin{tabular}{l c|c c c}
        & & \multicolumn{3}{c}{Images per class} \\
        Method & Teacher & 1 & 5 & 1\% \\
        \midrule
        \multicolumn{5}{l}{\small\bf\it Self-supervised baselines}\\
        DINO ViT-Ti & - & 18.6 & 37.8 & 47.0 \\
        DINO S/16 & - & 38.7 & 58.5 & 64.5 \\
        iBOT S/16 & - & 40.3 & 60.0 & 65.7 \\
        \midrule
        \multicolumn{5}{l}{\small\bf\it Self-supervised teachers}\\
        DINO S/8 & - & 45.3 & 65.1 & - \\
        iBOT L/16 & - & 47.5 & 65.9 & - \\
        MSN B/16 & - & 51.2 & 65.4 & - \\
        \midrule
        \multicolumn{5}{l}{\small\bf\it Self-supervised distillation}\\
        BINGO RN18 & SwAV RN50x2 & - & - & 48.2 \\
        BINGO RN18 & MoCoV2 RN152 & - & - & 50.3 \\
        RoB ViT-T/16 & DINO S/8 & \cc 23.7 & \cc 45.2 & \cc 54.0 \\
        RoB ViT-T/16 & iBOT L/16 & \cc 28.2 & \cc 48.6 & \cc 56.2 \\
        RoB ViT-T/16 & MSN B/16 & \cc \bf 40.9 & \cc \bf 54.5 & \cc \bf 58.7 \\
        \bottomrule
    \end{tabular}
    \caption{\textbf{RoB semi-supervised performance on ViT-Tiny}. MSN offers strong semi-supervised learning capabilities thanks to its strong in-distribution low-shot performance. RoB is able to transfer this low-shot capability to low-compute students.}
    \label{tab:msn_low_shot}
\end{table}

%% file: tables/sota.tex
\begin{table*}[!htb]
    \begin{minipage}{.5\linewidth}
      \centering
        \begin{tabular}{c|c c|c}
            Arch & Method & Teacher & IN1k \\
            \midrule
            \multicolumn{4}{l}{\small\bf\it Supervised pretraining}\\
            RN18 & Torchvision & - & 69.8 \\
            RN18 & RNSB A1 & - & 71.5 \\
            RN18 & DIST (KD) & RNSB50 A1 & 74.3 \\
            \midrule
            \multicolumn{4}{l}{\small\bf\it SSL without distillation}\\
            RN18 & SwAV & - & 59.5 \\
            RN18 & DINO & - & 62.2 \\
            RN18 & DINO+\cite{anonymous2023effective} 200 ep & - & 65.7 \\
            RN18 & DINO+\cite{anonymous2023effective} 400 ep & - & 66.8 \\
            \midrule
            \multicolumn{4}{l}{\small\bf\it SSL with distillation}\\
            RN18 & SimReg & SwAV RN50 & 65.8 \\
            RN18 & SimReg & BYOL RN50 & 66.8 \\
            RN18 & SimReg & Multi-teacher & \underline{67.0} \\
            RN18 & SEED & SwAV RN50x2 & 63.0 \\
            RN18 & CompRess & SwAV RN50 & 65.6 \\
            RN18 & DisCo & MoCoV2 R152 & 65.5 \\
            RN18 & BINGO & MoCoV2 R152 & 65.9 \\
            \midrule
            RN18 & RoB & SwAV RN50 & \cc 66.7 \\
            RN18 & RoB & SwAV RG128 & \cc 67.4 \\
            RN18 & RoB & DINO S/8 & \cc \bf 68.3 \\
            \midrule
            & $\Delta_\text{Best}$ & & \textcolor{darkgreen}{+1.3} \\
            \bottomrule
        \end{tabular}
    \end{minipage}%
    \begin{minipage}{.5\linewidth}
      \centering
        \begin{tabular}{c|c c|c}
            Arch & Method & Teacher & IN1k \\
            \midrule
            \multicolumn{4}{l}{\small\bf\it Supervised pretraining}\\
            RN34 & Torchvision & - & 73.3 \\
            RN34 & RNSB A1 & - & 76.3 \\
            RN34 & DIST (KD) & RNSB50 A1 & 77.8 \\
            \midrule
            \multicolumn{4}{l}{\small\bf\it SSL without distillation}\\
            RN34 & SwAV & - & 68.4  \\
            RN34 & DINO & - & 67.7 \\
            RN34 & DINO+\cite{anonymous2023effective} 200 ep & - & 69.7 \\
            RN34 & DINO+\cite{anonymous2023effective} 400 ep & - & \underline{70.8} \\
            \midrule
            \multicolumn{4}{l}{\small\bf\it SSL with distillation}\\
            RN34 & SimReg & SwAV RN50 & - \\
            RN34 & SimReg & SwAV RN50 & - \\
            RN34 & SimReg & Multi-teacher & - \\
            RN34 & SEED & SwAV RN50x2 & 65.7 \\
            RN34 & CompRess & SwAV RN50 & - \\
            RN34 & DisCo & MoCoV2 R152 & 68.1 \\
            RN34 & BINGO & MoCoV2 R152 & 69.1 \\
            \midrule
            RN34 & RoB & SwAV RN50 & \cc 70.4 \\
            RN34 & RoB & SwAV RG128 & \cc 71.3 \\
            RN34 & RoB & DINO/S8 & \cc \bf 72.6 \\
            \midrule
            & $\Delta_\text{Best}$ & & \textcolor{darkgreen}{+1.8} \\
            \bottomrule
        \end{tabular}
    \end{minipage} 
    \caption{\textbf{Comparison with the state of the art of self-supervision methods} with or without using KD on ImageNet top-1 accuracy, the de-facto common metric reported in all methods above. RoB improves on previous works by significant margins.}
    \label{tab:sota}
\end{table*}

%% file: tables/swav_mean.tex
\begin{table}[]
    \centering
    \begin{tabular}{l c|c c c}
        & & \multicolumn{3}{c}{Mean accuracy (6 tasks)} \\
        Method & Teacher & RN18 & RN34 & RN50 \\
        \midrule
        \multicolumn{2}{l}{\small\bf\it Methods without distillation}\\
        Torchvision & - & 56.23 & 57.17 & 60.93 \\
        RNSB A1 & - & 57.23 & 58.12 & 61.55 \\
        SwAV & - & 53.93 & 61.18 & 67.27 \\
        DINO & - & - & - & 67.08 \\
        \midrule
        \multicolumn{2}{l}{\small\bf\it Supervised distillation}\\
        DIST & RN50 SB A1 & 58.31 & 58.65 & - \\
        \midrule
        \multicolumn{2}{l}{\small\bf\it Self-supervised distillation}\\
        RoB & SwAV RN50 & \cc 60.35 & \cc 62.43 & \cc -  \\
        RoB & SwAV RG128 & \cc 60.45 & \cc 62.55 & \cc - \\
        RoB & DINO S/8 & \cc \bf 60.48 & \cc \bf 63.08 & \cc \bf 67.73 \\
        \midrule
        & $\Delta_\text{Best}$ & \textcolor{darkgreen}{+2.17} & \textcolor{darkgreen}{+1.90} & \textcolor{darkgreen}{+0.46} \\
        \bottomrule
    \end{tabular}
    \caption{\textbf{ResNet self-supervised students}. RoB applied to SwAV (teacher with same architecture as the student) or DINO (cross-architecture family distillation). Self-supervised distillation surpasses supervised pretraining on transfer tasks and competes with supervised distillation.}
    \label{tab:swav_mean}
\end{table}

%% file: tables/ablation_views.tex
\begin{table}[]
    \centering
    \begin{tabular}{c c|c c c}
        Method & Teacher & IN1K & IN1k 1\% & iNat18 \\
        \midrule
        RoB (300ep) & DINO S/8 & 71.4 & 54.0 & 39.8 \\
        \ac \textbf{w/o} multi-crop & DINO S/8 & 70.4 & 53.4 & 35.4 \\
        & $\Delta$ & -1.0 & -0.6 & -4.4 \\
        \midrule
        RoB (300ep) & DINO S/8 & 71.4 & 54.0 & 39.8 \\
        \ac w/i cross-view & DINO S/8 & 71.4 & 54.0 & 39.4 \\
        & $\Delta$ & -0.0 & -0.0 & -0.5 \\
        \midrule
        RoB (300ep) & iBOT L/16 & 71.8 & 56.2 & 37.4 \\
        \ac w/i cross-view & iBOT L/16 & 71.7 & 55.9 & 36.5 \\
        & $\Delta$ & -0.1 & -0.3 & -0.9 \\
        \bottomrule
    \end{tabular}
    \caption{\textbf{Multi-crop and identical-view assignment} are needed get the best results in-distribution and in transfer tasks. In particular on iNaturalist18, not using multi-crop leads to a significant $4.4\%$ performance drop, while the cross-view predictions accounts for up to $1\%$ performance drop on the same benchmark when compared to identical-view predictions.}
    \label{tab:ablation_views}
\end{table}

%% file: tables/ablation_head.tex
\begin{table}[]
    \centering
    \begin{tabular}{c c|c c c c}
        Method & Teacher & IN1K & iNat18 & Cifar100 \\
        \midrule
        \multicolumn{2}{l}{\small\bf\it Resnet18 student}\\
        RoB (300ep) & SwAV RN50 & 66.7 & \bf 37.2 & \bf 69.2 \\
        \ac MLP head & SwAV RN50 & \bf 66.8 & 36.0 & 67.8 \\
        & $\Delta$ & +0.1 & -1.2 & -1.4 \\
        \midrule
        \multicolumn{2}{l}{\small\bf\it ViT-Ti/16 student}\\
        RoB (300ep) & iBOT L/16 & 71.8 & \bf 37.4 & \bf 76.8 \\
        \ac MLP head & iBOT L/16 & \bf 72.0 & 36.1 & 71.9 \\
        & $\Delta$ & +0.2 & -1.6 & -4.9 \\
        \midrule
        \multicolumn{2}{l}{\small\bf\it ViT-Ti/16 student}\\
        RoB (300ep) & DINO S/8 & 71.4 & \bf 39.8 & \bf 77.1 \\
        \ac MLP head & DINO S/8 & \bf 71.7 & 38.2 & 72.7 \\
        & $\Delta$ & +0.3 & -1.6 & -5.0 \\
        \midrule
        \multicolumn{2}{l}{\small\bf\it Resnet18 student}\\
        RoB (100ep) & SwAV RN50 & 65.4 & 35.4 & 67.3 \\
        \ac Partial head & SwAV RN50 & 54.0 & 11.4 & 50.5 \\
        & $\Delta$ & -11.4 & -24.0 & -16.8 \\
        \bottomrule
    \end{tabular}
    \caption{\textbf{Student projection head designs.} Sticking to the projection head used in the joint-embedding approach being distilled seems best overall. Replacing the student head with a MLP decreases transfer performance, while trying to use the teacher prototypes in the student head leads to optimization difficulties.}
    \label{tab:ablation_head}
\end{table}

%% file: tables/appendix/mobile_net_transfer.tex
\begin{table*}[]
    \centering
    \begin{tabular}{c|c c |c c c c c c | c}
        Arch & Method & Teacher & IN1k & iNat18 & Clevr/C & Clevr/D & Cifar100 & Places & Mean \\
        \midrule
        MobileNetV3 & Torchvision~\cite{marcel2010torchvision} & - & 74.0 & 45.0 & 73.6 & 64.5 & 73.1 & 52.7 & 63.82 \\
        MobileNetV3 & timm~\cite{rw2019timm} & - & \bf 75.7 & \bf 47.8 & 73.9 & 62.0 & \bf 74.0 & 53.5 & 64.48 \\
        \midrule
        MobileNetV3 & RoB & SwAV RN50 & 70.3 & 45.0 & \bf 83.0 & \bf 69.5 & \bf 73.8 & \bf 54.7 & \cc \bf 66.05 \\
        \bottomrule
    \end{tabular}
    \caption{\textbf{MobileNetV3 transfer results}. When compared against supervised pretraining, RoB-SwAV trains MobileNetV3 student that surpass (in average on 6 tasks) the performance of supervised pretraining. This is similar to what we observed for architectures with ViT-Tiny, ResNet18 and ResNet34, despite MobileNetV3 having one order less compute.}
    \label{tab:mobile_nets}
\end{table*}

%% file: tables/appendix/mobile_net_sota.tex
\begin{table}[]
    \centering
    \begin{tabular}{c|c c |c }
        Arch & Method & Teacher & IN1k \\
        \midrule
        \multicolumn{4}{l}{\small\bf\it Supervised baselines}\\
        MobileNetV3 & Torchvision & - & 74.0 \\
        MobileNetV3 & timm~\cite{rw2019timm} & - & 75.7 \\
        \midrule
        \multicolumn{4}{l}{\small\bf\it Self-supervised methods}\\
        MobileNetV3 & DisCo & SwAV RN50x2 & 58.9 \\
        MobileNetV3 & DisCo & MoCoV2 RN-101 & 65.7 \\
        MobileNetV3 & SEED & MoCoV2 RN-152 & 61.4 \\
        MobileNetV3 & SEED & SwAV RN50x2 & \underline{68.2} \\
        MobileNetV3 & BINGO & SwAV RN50x2 & 67.1 \\
        \midrule
        MobileNetV3 & RoB & SwAV RN50 & \cc \bf 70.3 \\
        & & $\Delta_\text{Best}$ & \textcolor{darkgreen}{+2.1} \\
        \bottomrule
    \end{tabular}
    \caption{\textbf{MobileNetV3 results against the state of the art of self-supervision methods}, on ImageNet top-1 accuracy, the common metric reported in all methods above. RoB improves on previous work by significant margins, using with a smaller teacher.}
    \label{tab:mobile_nets_sota}
\end{table}

%% file: tables/appendix/knn.tex
\begin{table}[]
    \centering
    \begin{tabular}{c |l l | c c}
        Arch & Method & Teacher & 10NN & 20NN \\
        \midrule
        \multicolumn{5}{l}{\small\bf\it ResNet18 results}\\
        RN18 & CompRess & MoCoV2 RN50 & 53.5 & - \\
        RN18 & SEED & SwAV RN50x2 & 55.3 & - \\
        RN18 & BINGO & SwAV RN50x2 & 61.0 & - \\
        RN18 & RoB & SwAV RN50 & \cc 60.8 & \cc 61.1 \\
        RN18 & RoB & SwAV RG128 & \cc 61.9 & \cc 62.3 \\
        RN18 & RoB & DINO-S/8 & \cc \bf 63.7 & \cc \bf 63.9 \\
        \midrule
        \multicolumn{5}{l}{\small\bf\it ResNet34 results}\\
        RN34 & SEED & SwAV RN50x2 & 58.2 & - \\
        RN34 & BINGO & SwAV RN50x2 & 64.9 & - \\
        RN34 & RoB & SwAV RN50 & \cc 65.5 & \cc 65.7 \\
        RN34 & RoB & SwAV RG128 & \cc 67.0 & \cc 67.1 \\
        RN34 & RoB & DINO-S/8 & \cc \bf 68.6 & \cc \bf 68.7 \\
        \midrule
        \multicolumn{5}{l}{\small\bf\it ViT-Tiny results}\\
        ViT-Ti & RoB& DINO S/8 & \cc \bf 69.5 & \cc \bf 69.5 \\
        ViT-Ti & RoB& iBOT L/16 & \cc \bf 69.5 & \cc 69.4 \\
        \bottomrule
    \end{tabular}
    \caption{\textbf{Nearest Neighbor evaluations with RoB} on ImageNet, compared to the state of the art on kNN for self-supervised methods. RoB supasses other methods with significant margins.}
    \label{tab:knn_results}
    \vspace{1.0em}
\end{table}

%% file: tables/appendix/teachers_list.tex
\begin{table}[h]
    \centering
    \begin{tabular}{c|c c|c|c c}
        Teacher & Params & GFlops & Method & IN1k \\
        \midrule
        \multicolumn{4}{l}{\small\bf\it Supervised teachers}\\
        ResNet50 & 24M & 4.1 & RNSB A1 & 80.1 \\
        RegNet16Gf & 81M & 16.0 & DeiT & 82.9 \\
        \midrule
        \multicolumn{4}{l}{\small\bf\it Self-supervised teachers}\\
        ViT-S/8 & 22M & 22.4 & DINO & 79.7 \\
        ViT-B/16 & 86M & 17.6 & MSN & 77.2 \\
        ViT-B/16 & 86M & 17.6 & iBOT & 79.6 \\
        ViT-L/16 & 305M & 61.6 & iBOT & 81.3 \\
        ResNet50 & 24M & 4.1 & SwAV & 75.0 \\
        RegNet128Gf & 638M & 127.9 & SwAV & 78.9 \\
        \bottomrule
    \end{tabular}
    \caption{\textbf{List of the teachers}, supervised and self-supervised, used throughout the experiments. The supervised teachers are overall stronger on ImageNet.}
    \label{tab:teacher_list}
\end{table}

%% file: tables/appendix/student_list.tex
\begin{table}[h]
    \centering
    \begin{tabular}{c|c c| c}
        Teacher & Params & GFlops & Supervised IN1k\\
        \midrule
        MobileNet-V3 & 3M & 0.24 & 75.7 (timm) \\
        ViT-Tiny & 5.5M & 1.26 & 72.2 (DeiT) \\
        ResNet18 & 11M & 1.82 & 71.5 (RNSB A1) \\
        ResNet34 & 21M & 3.68 & 76.3 (RNSB A1) \\
        ResNet50 & 24M & 4.14 & 80.1 (RNSB A1) \\
        ViT-S/16 & 22M & 4.61 & 79.9 (DeiT) \\
        \bottomrule
    \end{tabular}
    \caption{\textbf{List of student architectures.} Parameters and GFlops are computed without the classifier head (encoder only). Supervised performance on IN1k is provided as reference.}
    \label{tab:student_list}
\end{table}

%% file: tables/appendix/resnets_full.tex
\begin{table*}[b]
    \centering
    \begin{tabular}{c|c c c|c c c c c c | c}
        Arch & Method & Teacher & Params & IN1K & iNat18 & Clevr/C & Clevr/D & Cifar100 & Places & Mean \\
        \midrule
        \multicolumn{10}{l}{\small\bf\it Methods without distillation}\\
        RN18 & Torchvision & - & - & 69.8 & 33.4 & 58.4 & 62.4 & 69.0 & 44.4 & 56.23 \\
        RN18 & RNSB A1 & - & - & 71.5 & 35.2 & 59.5 & 61.7 & 69.0 & 46.5 & 57.23 \\
        RN18 & SwAV & - & - & 59.5 & 25.5 & 68.6 & 63.8 & 60.8 & 45.4 & 53.93\\
        \midrule
        \multicolumn{10}{l}{\small\bf\it Supervised distillation}\\
        RN18 & DIST\_KD & RN50 SB A1 & 24M & \bf 74.3 & \bf 37.6 & 58.8 & 59.6 & \bf 72.2 & 47.4 & 58.31 \\
        \midrule
        \multicolumn{10}{l}{\small\bf\it Self-supervised distillation}\\
        RN18 & RoB & SwAV RN50 & 24M & 66.7 & 37.2 & \bf 73.0 & 66.5 & 69.2 & \bf 49.5 & \cc 60.35 \\
        RN18 & RoB & SwAV RG128 & 638M & 67.4 & 37.5 & 72.2 & \bf 67.3 & 69.0 & 49.3 & \cc 60.45 \\
        RN18 & RoB & DINO S/8 & 22M & 68.3 & \bf 41.0 & \bf 72.8 & 64.7 & 68.1 & 48.0 & \bf \cc 60.48 \\
        \bottomrule
    \end{tabular}
    \caption{\textbf{ResNet18 self-supervised students}. RoB applied to SwAV (teacher with same architecture as the student) or DINO (cross-architecture family distillation).}
    \label{tab:resnet18_full}
    \vspace{3.0em}
    
    \centering
    \begin{tabular}{c|c c c|c c c c c c | c}
        Arch & Method & Teacher & Params & IN1K & iNat18 & Clevr/C & Clevr/D & Cifar100 & Places & Mean \\
        \midrule
        \multicolumn{10}{l}{\small\bf\it Methods without distillation}\\
        RN34 & Torchvision & - & - & 73.3 & 34.5 & 55.7 & 62.2 & 70.8 & 46.5 & 57.17 \\
        RN34 & RNSB A1 & - & - & 76.3 & 36.2 & 56.8 & 59.9 & 71.4 & 48.1 & 58.12 \\
        RN34 & SwAV & - & - & 68.4 & 37.3 & 73.9 & 67.3 & 69.6 & 50.6 & 61.18 \\
        \midrule
        \multicolumn{10}{l}{\small\bf\it Supervised distillation}\\
        RN34 & DIST\_KD & RN50 SB A1 & 24M & \bf 77.8 & 36.6 & 56.2 & 58.6 & \bf 73.9 & 48.8 & 58.65 \\
        \midrule
        \multicolumn{10}{l}{\small\bf\it Self-supervised distillation}\\
        RN34 & RoB & SwAV RN50 & 24M & 70.4 & 39.5 & 73.6 & \bf 66.3 & 73.7 & 51.1 & \cc 62.43\\
        RN34 & RoB & SwAV RG128 & 638M & 71.3 & 40.3 & 72.6 & 65.9 & \bf 73.9 & \bf 51.3 & \cc 62.55 \\
        RN34 & RoB & DINO S/8 & 22M & 72.6 & \bf 43.7 & \bf 74.2 & 65.1 & 72.1 & 50.8 & \bf \cc 63.08 \\
        \bottomrule
    \end{tabular}
    \caption{\textbf{ResNet34 self-supervised students}. RoB applied to SwAV (teacher with same architecture as the student) or DINO (cross-architecture family distillation).}
    \label{tab:resnet34_full}
    \vspace{3.0em}
    
    \centering
    \begin{tabular}{c|c c c|c c c c c c | c}
        Arch & Method & Teacher & Params & IN1K & iNat18 & Clevr/C & Clevr/D & Cifar100 & Places & Mean \\
        \midrule
        \multicolumn{10}{l}{\small\bf\it Methods without distillation}\\
        RN50 & Torchvision & - & - & 76.1 & 39.0 & 66.4 & 63.2 & 71.2 & 49.7 & 60.93 \\
        RN50 & RNSB A1 & - & - & 80.1 & 43.6 & 62.6 & 55.9 & 73.7 & 53.4 & 61.55 \\
        RN50 & SwAV & - & - & 75.0 & 47.4 & 80.4 & 67.6 & 77.1 & 56.1 & 67.27 \\
        RN50 & DINO & - & - & 75.2 & 50.1 & 82.1 & 64.6 & 74.4 & 56.1 & 67.08 \\
        \midrule
        \multicolumn{10}{l}{\small\bf\it Self-supervised distillation}\\
        RN50 & RoB (100ep) & DINO S/8 & 22M & 76.6 & 52.1 & 80.6 & 66.6 & 75.7 & 54.8 & \bf \cc 67.73 \\
        \bottomrule
    \end{tabular}
    \caption{\textbf{ResNet50 self-supervised students}. RoB applied to a DINO-S/8 to produce a ResNet50 that is stronger than if trained with DINO without a teacher.}
    \label{tab:resnet50_full}
\end{table*}

%% file: main.bbl
\begin{thebibliography}{10}\itemsep=-1pt

\bibitem{anonymous2023effective}
Anonymous.
\newblock Effective self-supervised pre-training on low-compute networks
  without distillation.
\newblock In {\em Submitted to The Eleventh International Conference on
  Learning Representations}, 2023.
\newblock under review.

\bibitem{assran2022masked}
Mahmoud Assran, Mathilde Caron, Ishan Misra, Piotr Bojanowski, Florian Bordes,
  Pascal Vincent, Armand Joulin, Michael Rabbat, and Nicolas Ballas.
\newblock Masked siamese networks for label-efficient learning.
\newblock {\em arXiv preprint arXiv:2204.07141}, 2022.

\bibitem{bao2021beit}
Hangbo Bao, Li Dong, and Furu Wei.
\newblock Beit: Bert pre-training of image transformers.
\newblock {\em arXiv preprint arXiv:2106.08254}, 2021.

\bibitem{bardes2021vicreg}
Adrien Bardes, Jean Ponce, and Yann LeCun.
\newblock Vicreg: Variance-invariance-covariance regularization for
  self-supervised learning.
\newblock {\em arXiv preprint arXiv:2105.04906}, 2021.

\bibitem{https://doi.org/10.48550/arxiv.2106.05237}
Lucas Beyer, Xiaohua Zhai, Amélie Royer, Larisa Markeeva, Rohan Anil, and
  Alexander Kolesnikov.
\newblock Knowledge distillation: A good teacher is patient and consistent,
  2021.

\bibitem{bordes2022high}
Florian Bordes, Randall Balestriero, and Pascal Vincent.
\newblock High fidelity visualization of what your self-supervised
  representation knows about.
\newblock {\em Transactions on Machine Learning Research}, 2022.

\bibitem{bromley1993signature}
Jane Bromley, James~W Bentz, L{\'e}on Bottou, Isabelle Guyon, Yann LeCun, Cliff
  Moore, Eduard S{\"a}ckinger, and Roopak Shah.
\newblock Signature verification using a “siamese” time delay neural
  network.
\newblock {\em International Journal of Pattern Recognition and Artificial
  Intelligence}, 7(04):669--688, 1993.

\bibitem{caron2020unsupervised}
Mathilde Caron, Ishan Misra, Julien Mairal, Priya Goyal, Piotr Bojanowski, and
  Armand Joulin.
\newblock Unsupervised learning of visual features by contrasting cluster
  assignments.
\newblock {\em arXiv preprint arXiv:2006.09882}, 2020.

\bibitem{caron2021emerging}
Mathilde Caron, Hugo Touvron, Ishan Misra, Herv{\'e} J{\'e}gou, Julien Mairal,
  Piotr Bojanowski, and Armand Joulin.
\newblock Emerging properties in self-supervised vision transformers.
\newblock {\em arXiv preprint arXiv:2104.14294}, 2021.

\bibitem{chen2020simple}
Ting Chen, Simon Kornblith, Mohammad Norouzi, and Geoffrey Hinton.
\newblock A simple framework for contrastive learning of visual
  representations.
\newblock {\em preprint arXiv:2002.05709}, 2020.

\bibitem{chen2020mocov2}
Xinlei Chen, Haoqi Fan, Ross Girshick, and Kaiming He.
\newblock Improved baselines with momentum contrastive learning.
\newblock {\em arXiv preprint arXiv:2003.04297}, 2020.

\bibitem{dosovitskiy2020image}
Alexey Dosovitskiy, Lucas Beyer, Alexander Kolesnikov, Dirk Weissenborn,
  Xiaohua Zhai, Thomas Unterthiner, Mostafa Dehghani, Matthias Minderer, Georg
  Heigold, Sylvain Gelly, et~al.
\newblock An image is worth 16x16 words: Transformers for image recognition at
  scale.
\newblock {\em arXiv preprint arXiv:2010.11929}, 2020.

\bibitem{seed}
Zhiyuan Fang, Jianfeng Wang, Lijuan Wang, Lei Zhang, Yezhou Yang, and Zicheng
  Liu.
\newblock Seed: Self-supervised distillation for visual representation, 2021.

\bibitem{disco}
Yuting Gao, Jia-Xin Zhuang, Shaohui Lin, Hao Cheng, Xing Sun, Ke Li, and
  Chunhua Shen.
\newblock Disco: Remedy self-supervised learning on lightweight models with
  distilled contrastive learning, 2021.

\bibitem{https://doi.org/10.48550/arxiv.2103.01988}
Priya Goyal, Mathilde Caron, Benjamin Lefaudeux, Min Xu, Pengchao Wang, Vivek
  Pai, Mannat Singh, Vitaliy Liptchinsky, Ishan Misra, Armand Joulin, and Piotr
  Bojanowski.
\newblock Self-supervised pretraining of visual features in the wild, 2021.

\bibitem{goyal2021vissl}
Priya Goyal, Quentin Duval, Jeremy Reizenstein, Matthew Leavitt, Min Xu,
  Benjamin Lefaudeux, Mannat Singh, Vinicius Reis, Mathilde Caron, Piotr
  Bojanowski, Armand Joulin, and Ishan Misra.
\newblock Vissl.
\newblock \url{https://github.com/facebookresearch/vissl}, 2021.

\bibitem{goyal2022vision}
Priya Goyal, Quentin Duval, Isaac Seessel, Mathilde Caron, Mannat Singh, Ishan
  Misra, Levent Sagun, Armand Joulin, and Piotr Bojanowski.
\newblock Vision models are more robust and fair when pretrained on uncurated
  images without supervision.
\newblock {\em arXiv preprint arXiv:2202.08360}, 2022.

\bibitem{grill2020bootstrap}
Jean-Bastien Grill, Florian Strub, Florent Altch{\'e}, Corentin Tallec,
  Pierre~H Richemond, Elena Buchatskaya, Carl Doersch, Bernardo~Avila Pires,
  Zhaohan~Daniel Guo, Mohammad~Gheshlaghi Azar, et~al.
\newblock Bootstrap your own latent: A new approach to self-supervised
  learning.
\newblock {\em arXiv preprint arXiv:2006.07733}, 2020.

\bibitem{https://doi.org/10.48550/arxiv.2006.07733}
Jean-Bastien Grill, Florian Strub, Florent Altché, Corentin Tallec, Pierre~H.
  Richemond, Elena Buchatskaya, Carl Doersch, Bernardo~Avila Pires,
  Zhaohan~Daniel Guo, Mohammad~Gheshlaghi Azar, Bilal Piot, Koray Kavukcuoglu,
  Rémi Munos, and Michal Valko.
\newblock Bootstrap your own latent: A new approach to self-supervised
  learning, 2020.

\bibitem{he2021masked}
Kaiming He, Xinlei Chen, Saining Xie, Yanghao Li, Piotr Doll{\'a}r, and Ross
  Girshick.
\newblock Masked autoencoders are scalable vision learners.
\newblock {\em arXiv preprint arXiv:2111.06377}, 2021.

\bibitem{he2019moco}
Kaiming He, Haoqi Fan, Yuxin Wu, Saining Xie, and Ross Girshick.
\newblock Momentum contrast for unsupervised visual representation learning.
\newblock {\em arXiv preprint arXiv:1911.05722}, 2019.

\bibitem{he2016deep}
Kaiming He, Xiangyu Zhang, Shaoqing Ren, and Jian Sun.
\newblock Deep residual learning for image recognition.
\newblock In {\em Proceedings of the IEEE Conference on Computer Vision and
  Pattern Recognition}, pages 770--778, 2016.

\bibitem{hinton_knowledge_distillation}
Geoffrey Hinton, Oriol Vinyals, and Jeff Dean.
\newblock Distilling the knowledge in a neural network, 2015.

\bibitem{mobilenet}
Andrew Howard, Mark Sandler, Grace Chu, Liang-Chieh Chen, Bo Chen, Mingxing
  Tan, Weijun Wang, Yukun Zhu, Ruoming Pang, Vijay Vasudevan, Quoc~V. Le, and
  Hartwig Adam.
\newblock Searching for mobilenetv3, 2019.

\bibitem{https://doi.org/10.48550/arxiv.2205.10536}
Tao Huang, Shan You, Fei Wang, Chen Qian, and Chang Xu.
\newblock Knowledge distillation from a stronger teacher, 2022.

\bibitem{johnson2017clevr}
Justin Johnson, Bharath Hariharan, Laurens Van Der~Maaten, Li Fei-Fei, C
  Lawrence~Zitnick, and Ross Girshick.
\newblock Clevr: A diagnostic dataset for compositional language and elementary
  visual reasoning.
\newblock In {\em Proceedings of the IEEE conference on computer vision and
  pattern recognition}, pages 2901--2910, 2017.

\bibitem{compress}
Soroush~Abbasi Koohpayegani, Ajinkya Tejankar, and Hamed Pirsiavash.
\newblock Compress: Self-supervised learning by compressing representations,
  2020.

\bibitem{krizhevsky2009learning}
Alex Krizhevsky, Geoffrey Hinton, et~al.
\newblock Learning multiple layers of features from tiny images.
\newblock 2009.

\bibitem{marcel2010torchvision}
S{\'e}bastien Marcel and Yann Rodriguez.
\newblock Torchvision the machine-vision package of torch.
\newblock In {\em Proceedings of the 18th ACM international conference on
  Multimedia}, pages 1485--1488, 2010.

\bibitem{DBLP:journals/corr/abs-2201-05131}
K.~L. Navaneet, Soroush~Abbasi Koohpayegani, Ajinkya Tejankar, and Hamed
  Pirsiavash.
\newblock Simreg: Regression as a simple yet effective tool for self-supervised
  knowledge distillation.
\newblock {\em CoRR}, abs/2201.05131, 2022.

\bibitem{pytorch}
Adam Paszke, Sam Gross, Francisco Massa, Adam Lerer, James Bradbury, Gregory
  Chanan, Trevor Killeen, Zeming Lin, Natalia Gimelshein, Luca Antiga, Alban
  Desmaison, Andreas Kopf, Edward Yang, Zachary DeVito, Martin Raison, Alykhan
  Tejani, Sasank Chilamkurthy, Benoit Steiner, Lu Fang, Junjie Bai, and Soumith
  Chintala.
\newblock Pytorch: An imperative style, high-performance deep learning library.
\newblock In {\em Advances in Neural Information Processing Systems 32}, pages
  8024--8035. Curran Associates, Inc., 2019.

\bibitem{scikit-learn}
F. Pedregosa, G. Varoquaux, A. Gramfort, V. Michel, B. Thirion, O. Grisel, M.
  Blondel, P. Prettenhofer, R. Weiss, V. Dubourg, J. Vanderplas, A. Passos, D.
  Cournapeau, M. Brucher, M. Perrot, and E. Duchesnay.
\newblock Scikit-learn: Machine learning in {P}ython.
\newblock {\em Journal of Machine Learning Research}, 12:2825--2830, 2011.

\bibitem{https://doi.org/10.48550/arxiv.1412.6550}
Adriana Romero, Nicolas Ballas, Samira~Ebrahimi Kahou, Antoine Chassang, Carlo
  Gatta, and Yoshua Bengio.
\newblock Fitnets: Hints for thin deep nets, 2014.

\bibitem{russakovsky2015imagenet}
Olga Russakovsky, Jia Deng, Hao Su, Jonathan Krause, Sanjeev Satheesh, Sean Ma,
  Zhiheng Huang, Andrej Karpathy, Aditya Khosla, Michael Bernstein,
  Alexander~C. Berg, and Li Fei-Fei.
\newblock Imagenet large scale visual recognition challenge.
\newblock {\em International Journal of Computer Vision}, 115(3):211--252,
  2015.

\bibitem{https://doi.org/10.48550/arxiv.2107.14762}
Haizhou Shi, Youcai Zhang, Siliang Tang, Wenjie Zhu, Yaqian Li, Yandong Guo,
  and Yueting Zhuang.
\newblock On the efficacy of small self-supervised contrastive models without
  distillation signals, 2021.

\bibitem{Tian2020Contrastive}
Yonglong Tian, Dilip Krishnan, and Phillip Isola.
\newblock Contrastive representation distillation.
\newblock In {\em International Conference on Learning Representations}, 2020.

\bibitem{touvron2021training}
Hugo Touvron, Matthieu Cord, Matthijs Douze, Francisco Massa, Alexandre
  Sablayrolles, and Herv{\'e} J{\'e}gou.
\newblock Training data-efficient image transformers \& distillation through
  attention.
\newblock In {\em International Conference on Machine Learning}, pages
  10347--10357. PMLR, 2021.

\bibitem{van2018inaturalist}
Grant Van~Horn, Oisin Mac~Aodha, Yang Song, Yin Cui, Chen Sun, Alex Shepard,
  Hartwig Adam, Pietro Perona, and Serge Belongie.
\newblock The inaturalist species classification and detection dataset.
\newblock In {\em Proceedings of the IEEE conference on computer vision and
  pattern recognition}, pages 8769--8778, 2018.

\bibitem{rw2019timm}
Ross Wightman.
\newblock Pytorch image models.
\newblock \url{https://github.com/rwightman/pytorch-image-models}, 2019.

\bibitem{DBLP:journals/corr/abs-2110-00476}
Ross Wightman, Hugo Touvron, and Herv{\'{e}} J{\'{e}}gou.
\newblock Resnet strikes back: An improved training procedure in timm.
\newblock {\em CoRR}, abs/2110.00476, 2021.

\bibitem{https://doi.org/10.48550/arxiv.2110.00476}
Ross Wightman, Hugo Touvron, and Hervé Jégou.
\newblock Resnet strikes back: An improved training procedure in timm, 2021.

\bibitem{xu2022bag}
Haohang Xu, Jiemin Fang, XIAOPENG ZHANG, Lingxi Xie, Xinggang Wang, Wenrui Dai,
  Hongkai Xiong, and Qi Tian.
\newblock Bag of instances aggregation boosts self-supervised distillation.
\newblock In {\em International Conference on Learning Representations}, 2022.

\bibitem{zbontar2021barlow}
Jure Zbontar, Li Jing, Ishan Misra, Yann LeCun, and St{\'e}phane Deny.
\newblock Barlow twins: Self-supervised learning via redundancy reduction.
\newblock {\em arXiv preprint arXiv:2103.03230}, 2021.

\bibitem{https://doi.org/10.48550/arxiv.1910.04867}
Xiaohua Zhai, Joan Puigcerver, Alexander Kolesnikov, Pierre Ruyssen, Carlos
  Riquelme, Mario Lucic, Josip Djolonga, Andre~Susano Pinto, Maxim Neumann,
  Alexey Dosovitskiy, Lucas Beyer, Olivier Bachem, Michael Tschannen, Marcin
  Michalski, Olivier Bousquet, Sylvain Gelly, and Neil Houlsby.
\newblock A large-scale study of representation learning with the visual task
  adaptation benchmark, 2019.

\bibitem{zhou2014learning}
Bolei Zhou, Agata Lapedriza, Jianxiong Xiao, Antonio Torralba, and Aude Oliva.
\newblock Learning deep features for scene recognition using places database.
\newblock {\em Advances in neural information processing systems}, 27, 2014.

\bibitem{zhou2021ibotyes}
Jinghao Zhou, Chen Wei, Huiyu Wang, Wei Shen, Cihang Xie, Alan Yuille, and Tao
  Kong.
\newblock Ibot: Image bert pre-training with online tokenizer.
\newblock {\em arXiv preprint arXiv:2111.07832}, 2021.

\end{thebibliography}
